\documentclass[10pt, journal]{IEEEtran}

\usepackage{graphicx}  
\usepackage{cite}
\usepackage{amsthm}
\usepackage[ruled,vlined]{algorithm2e}
\usepackage{amsmath}   
\usepackage{amssymb}   
\usepackage{hyperref}
\usepackage[dvipsnames]{xcolor}
\usepackage{array}
\usepackage{multirow}
\usepackage{comment}
\renewcommand{\arraystretch}{1.2}
\usepackage{subcaption}
\newcommand{\vect}[1]{\boldsymbol{#1}}
\newtheorem{theorem}{Theorem}

\newtheorem{problem}{Problem}

\title{Onboard Ranging-based Relative Localization and Stability for Lightweight Aerial Swarms (Extended Version)}
\author{Shushuai Li, Feng Shan, Jiangpeng Liu, Mario Coppola, Christophe de Wagter and Guido C. H. E. de Croon
   \thanks{Shushuai Li, Mario Coppola, Christophe de Wagter and Guido C. H. E. de Croon are with the Faculty of Aerospace Engineering,
       Delft University of Technology, 2629 HS Delft, The Netherlands (e-mail: s.li-6@tudelft.nl; m.coppola@tudelft.nl; c.deWagter@tudelft.nl; g.c.h.e.decroon@tudelft.nl).
       Feng Shan is with the School of Computer Science and Engineering, while Jiangpeng Liu is with the School of Software Engineering, Southeast University, Nanjing, Jiangsu 210096, China(e-mail: shanfeng@seu.edu.cn; liujiangpeng@seu.edu.cn)
   }%
}

\begin{document}

\maketitle

\begin{abstract}

Lightweight aerial swarms have potential applications in scenarios where larger drones fail to operate efficiently. 
The primary foundation for lightweight aerial swarms is \emph{efficient relative localization}, which enables cooperation and collision avoidance. 
Computing the real-time position is challenging due to extreme resource constraints. 
This paper presents an autonomous relative localization technique for lightweight aerial swarms without infrastructure by fusing ultra-wideband wireless distance measurements and the shared state information (e.g., velocity, yaw rate, height) from neighbors. 
This is the first fully autonomous, tiny, fast, and accurate relative localization scheme implemented on a team of 13 lightweight (33 grams) and resource-constrained (168MHz MCU with 192 KB memory) aerial vehicles.
The proposed resource-constrained swarm ranging protocol is scalable, and a surprising theoretical result is discovered: the unobservability poses no issues because the state drift leads to control actions that make the state observable again. 
By experiment, less than 0.2m position error is achieved at the frequency of 16Hz for as many as 13 drones. 
The code is open-sourced, and the proposed technique is relevant not only for tiny drones but can be readily applied to many other resource-restricted robots.
Video and code can be found at \textnormal{\url{https://shushuai3.github.io/autonomous-swarm/}}.
\end{abstract}




\section{Introduction}

With rapid progress in electronic manufacturing, increasingly lightweight aerial vehicles are becoming a part of our daily lives.
These lightweight aerial vehicles and swarms ~\cite{laghari2023unmanned} have found unique applications in scenarios where larger drones fail to operate efficiently due to size constraints and higher risk factors.
Examples include conducting cooperative rescue operations within buildings, 
cooperative mapping in narrow spaces, 
cooperative exploration in unknown environments and closely following humans.

A primary foundation for lightweight aerial swarms is \emph{efficient relative localization}, which enables collaborative cooperation and collision avoidance. 
However, how to compute the real-time position within the swarm is quite challenging, because lightweight aerial vehicles are extremely constrained in resources. 
For example, a typical off-the-shelf lightweight aerial vehicle, Crazyflie 2.1 from Bitcraze, is less than 10 centimeters and weights only 33 grams, 
and it carries a 168 MHz microcontroller unit (MCU) with only 192 KB memory, extremely constrained on both computation and memory resources.

Recent advancements indicate that wireless ranging-based relative positioning in lightweight aerial swarms is a promising approach.
Nevertheless, several significant challenges remain to be addressed.
\underline{First}, there are no existing ranging protocols that support a large number of vehicles to range simultaneously for resource-constrained large-scale swarms. 
\underline{Second}, traditional methods necessitate prior knowledge of the vehicle's location, a requirement that is impractical for many lightweight aerial vehicle applications.
\underline{Third}, there are considerable parts of the relative state space that are \emph{unobservable}, meaning that the filter may not converge to the true state. How the unobservable part affects the localization is unknown.


In this paper, we propose a novel onboard Ultra-Wideband (UWB) wireless ranging-based autonomous relative localization technique, for which we study the challenges mentioned above.
\textbf{Contributions} involve: 
1) the first fully autonomous, tiny, fast, and accurate relative localization scheme implemented on a team of 13 lightweight (33 grams) and resource-constrained (168MHz MCU with 192 KB memory) aerial vehicles, and they perform fully onboard relative localization in an indoor environment; 
2) the proposed resource-constrained ranging protocol can enable large-scale swarms to perform fast and stable communication and ranging, \emph{e.g.}, $16$ Hz ranging communication for 13 aerial vehicles and is scalable;
3) a surprising theoretical result is discovered and validated: the unobservability poses no issues because the state drift and sensory noise lead to control actions that make the state observable again, and this \emph{self-regulated estimation convergence} is validated by simulation and real-world experiments. 
4) public release of the code to the community. It can be run on off-the-shelf Crazyflie quadrotors by peers. 

\section{RELATED WORK}
Localization methods have been widely studied due to their critical importance in enabling accurate positioning and coordination among autonomous agents.
Traditionally, the localization method employed for conventional aerial vehicles has been dependent on external positioning systems.
In \cite{vasarhelyi2018optimized}, 30 drones exhibit outdoor flocking behavior by relying on a Global Navigation Satellite System (GNSS) for positioning.
However, lightweight aerial vehicles often operate in indoor or confined areas where it is impossible to receive signals from external positioning systems like GNSS.
Alternatively, some indoor positioning systems rely on beacons or anchors to provide localization information for aerial swarms \cite{oumar2023indoor, jia2022composite, brandstatter2023multi}.
However, these external systems need to be deployed in advance, which is impractical in dangerous or unknown areas.

Another method relies on vision. 
Simple patterns are designed in ~\cite{li2022self, bonato2023ultra,crupi2024high} that could be detected by a monocular camera to calculate the relative position of other vehicles. 
Some more recent techniques involve detecting active LED tags \cite{stuckey2023real} and ultraviolet light \cite{horyna2022uvdar}.
However, visual positioning systems depend heavily on good lighting and a clear view to work precisely.
In places with low light or changing lighting conditions, like indoors or at night, their accuracy can drop significantly. 
Also, these systems need to process a lot of visual information, which is tough for computation and memory-limited lightweight aerial vehicles.
Given these issues, using visual positioning for lightweight aerial vehicles in various situations seems unfeasible, pointing to a need for solutions that are more flexible and use less onboard resources.

As an alternative approach, relative localization based on wireless communication between drones has the advantage of being lightweight and suitable for resource-constrained aerial vehicles.
The aerial robots use wireless antennas to exchange state information (e.g., velocity, yaw rate, height) and combine these with relative range measurements obtained from the antennas, which attracts recent attensions~\cite{guo2017relativelocalization, nguyen2023relative, cao2021relative}.
Guo \textit{et al.} \cite{guo2017relativelocalization}  puts forward an indirect cooperative relative localization method based on stationary bacon, estimating a robot's position relative to its neighbors using only distance and self-displacement measurements.
In \cite{nguyen2023relative}, optimization-based solutions are presented, including a QCQP approach and its SDP relaxation. While both methods achieve high localization accuracy on onboard computers, they require approximately 15 ms for localization, posing challenges for lightweight drones.
Cao \textit{et al.} \cite{cao2021relative} propose a method for estimating relative pose among robots using multiple UWB ranging nodes, necessitating larger robots and significant computing power.
Consequently, effective localization methods for lightweight drones remain a substantial challenge.

Communication and ranging are the foundational elements of swarming.
The common double-sided two-way ranging (DS-TWR) protocol is designed for the one-to-one ranging. 
A simple extension of the DS-TWR protocol has been proposed \cite{bux2005chapter} to deal with many-to-many ranging operations, that use the token ring technique to control the ranging process.
While this protocol considerably enhances the many-to-many ranging effect, it still has several limitations.
Such as whenever two sides are exchanging messages, these messages can be heard by a number of neighbors because of the broadcast nature of the wireless communication. 
However, they are ignored, which makes the protocol inefficient. 

The newly amended IEEE standard 802.15.4z-2020 \cite{ieee802154} also includes many-to-many ranging protocol based on the DS-TWR protocol. In this protocol, the basic time unit is called ranging round, and the swarm ranging distances are updated once in every ranging round. 
Although this many-to-many ranging protocol from the new IEEE standard is a good extension to the DS-TWR, it lacks
support for dense networks.
A ranging round may take a very long time to complete for dense networks, because a device or robot may have many neighbors to ranging with, and each of them must be allocated multiple time slots.
Therefore, the denser a network is, the longer a ranging round will take.
Overall, there are no perfect ranging protocols that support a large number of vehicles to range simultaneously for resource-constrained large-scale swarms. 
\section{System Definition and Problem Formulation}
\label{sec:uwb_system}

\subsection{System Definition}

For clarity of analysis, the model of two arbitrary robots is discussed here, in which robot $i$ needs to estimate the relative position of another robot $\{j\mid j\in\mathbb{N}, j\neq i\}$, where $\mathbb{N}=\{1,2,...,N \}$, and $N$ is the number of robots.

\begin{figure}[h]
    \centering
    \includegraphics[width=0.47\textwidth]{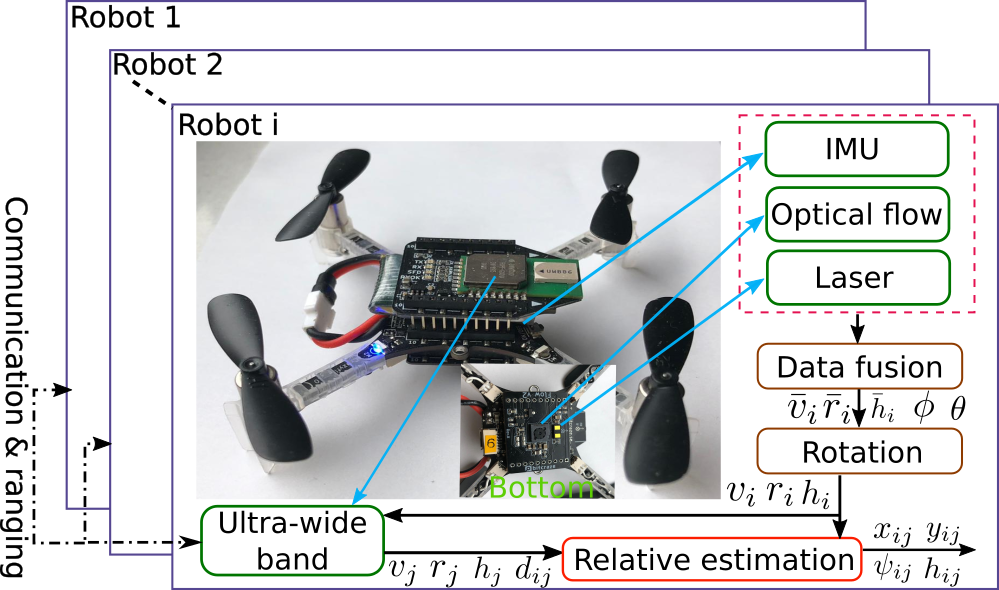}
    \caption{The scheme of the multi-robot system and all onboard sensors. Specifically, each robot has an inertial measurement unit (IMU), an optical flow sensor, and a downward-pointing laser sensor for obtaining acceleration, rotation rates, velocities, and height. This information is fused by an onboard filter to get the body-frame velocity, yaw rate, and height, which is further rotated to get the horizontal-frame velocity, yaw rate and height. By UWB wireless ranging and communication, other robots' state information is received and combined, the relative positions and yaw are estimated.}
    \label{fig:uwb_sensor}
\end{figure}

Fig.~\ref{fig:uwb_sensor} shows how the onboard sensing data is processed and how multiple robots communicate and range with each other, aiming at relative estimation. 
For each aerial robot, the 3-axis velocity $\bar{\vect{v}}=[\bar{v}^x, \bar{v}^y, \bar{v}^z]^T$, pitch $\theta$, and roll $\phi$ in the body frame can be obtained by fusing IMU, height, and optical flow measurements.
The yaw rate $\bar{r}$ in the body frame is provided by a gyroscope. 
The range $d_{ij}$, meaning the distance between robots $i$ and $j$, can be measured by UWB sensors. The variable $h$ is the vertical height calculated from a downward laser measurement $\bar{h}_i$. The final output $x_{ij}$, $y_{ij}$, and $h_{ij}$ denote the relative position between the $i^{\rm{th}}$ and $j^{\rm{th}}$ robots.

\subsection{Problem Formulation}

\begin{figure}[ht]
    \centering
    \includegraphics[width=0.47\textwidth]{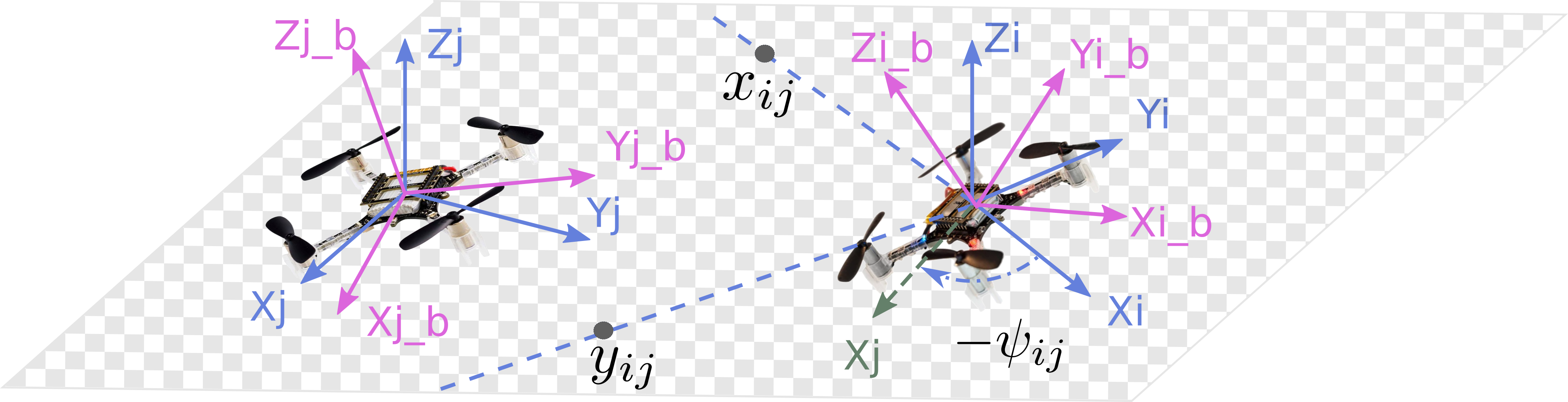}
    \caption{The diagram of the relative kinematic model, composed by two robots shown in a horizontal plane for simplicity (as they can be at different heights with the relative height $h_{ij}$). 3D purple axes represent the body frame of each robot, while the 3D blue axes denote the horizontal frame with a vertical z-axis. The relative 2D position $[x_{ij}, y_{ij}]$ and relative yaw $\psi_{ij}$ of $j^{\mathrm{th}}$ robot is shown in $i^{\mathrm{th}}$ robot horizontal frame in this figure.}
    \label{fig:uwb_model}
\end{figure}

Each drone localizes other drones in an ego-centered \emph{horizontal frame} with a purely vertical z-axis, as shown in blue lines in Fig \ref{fig:uwb_model}.

The 2-axis velocity $\vect{v}=[v^x, v^y]^T$ in the horizontal frame for each robot can be obtained from body-frame velocity $\vect{\bar{v}}$ and the attitude based on the rotation matrix:
\begin{equation} \label{eq:uwb_vel_body2hori}
  \vect{v} = 
  \begin{bmatrix} c(\theta) & 0 & s(\theta) \\ s(\phi)s(\theta) & c(\phi) & -c(\theta)s(\phi) \end{bmatrix}
  \vect{\bar{v}},
\end{equation}
where $s(\cdot)$, $c(\cdot)$ and $t(\cdot)$ denote $sin(\cdot)$, $cos(\cdot)$ and $tan(\cdot)$, respectively. In addition, the yaw rate $r$ in horizontal frame can be calculated with the gyroscope measurements $\bar{p}$ and $\bar{r}$:
$  r = -s(\theta)/c(\phi)\bar{p} + c(\theta)/c(\phi)\bar{r}.$

Given an arbitrary pair of robots $i$ and $j$, as shown in Fig. \ref{fig:uwb_model}, the inputs for estimation are represented by $\vect{U}_{ij}=[\vect{v}_i^T, r_i, \vect{v}_j^T, r_j]^T$. The measurements consist of $h_i$, $h_j$ and $d_{ij}$.
Define the relative state of $j^{\mathrm{th}}$ robot in the $i^{\mathrm{th}}$ robot's horizontal frame as $\vect{X}_{ij}=[x_{ij}, y_{ij}, \psi_{ij}]^T$, which is the core problem of this paper and needs to be estimated based on the inputs and measurements.

The kinematic model of the swarm of aerial robots can be derived based on Newton's formulas, which is given as
\begin{equation} \label{eq:uwb_model}
  \dot{\vect{X}}_{ij}=f(\vect{X}_{ij},\vect{U}_{ij})=
  \begin{bmatrix} R(\psi_{ij})\vect{v}_j-\vect{v}_i-Sr_i\vect{p}_{ij} \\ r_j-r_i \end{bmatrix},
\end{equation}
where $\vect{v}_i=[v_i^x, v_i^y]^T$ and $\vect{v}_j=[v_j^x, v_j^y]^T$ represent the 2-axis horizontal velocity of two robots; $\vect{p}_{ij}=[x_{ij}, y_{ij}]^T$ is a part of the relative state $\vect{X}_{ij}$ meaning 2-axis relative position. $R(\cdot)=[c(\cdot),-s(\cdot);s(\cdot),c(\cdot)]$ is the rotation function from $j^{\mathrm{th}}$ horizontal frame to $i^{\mathrm{th}}$ horizontal frame, and $S=[0,-1;1,0]$.

\section{Communication and Ranging for Relative Localization}
\label{sec:uwb_comm}


An efficient UWB wireless ranging protocol has been developed to facilitate simultaneous wireless communication and swarm ranging. 
The proposed protocol enables robots to accurately compute distances to all neighboring peers concurrently.

The DS-TWR protocol~\cite{ieee802154z} forms the basis of our approach, involving the exchange of four message types in a specific order between two sides. The sequence begins with one side initiating a \emph{poll} message, followed by alternating \emph{response}, \emph{final}, and \emph{report} messages.

By defining the transmission and reception timestamps as $T_p$, $R_p$, $T_r$, $R_r$, $T_f$, and $R_f$ respectively,
we introduce the concepts of reply time duration and round time duration for both sides, $a_d=R_{r}-T_{p}, b_p=T_{r}-R_{p}, b_d=R_{f}-T_{r}, a_p=T_{f}-R_{r}$.

Let $t_p$ represent the time of flight (ToF), which corresponds to the propagation time of the radio signal, which can be calculated by 
$t_p=\frac{a_d b_d-a_p b_p}{a_d+b_d+a_p+b_p}$~\cite{ieee802154z}.
Then, the distance can be estimated by the ToF.

\begin{figure}[b]
    \centering

    \begin{subfigure}[b]{0.23\textwidth}
        \centering
        \includegraphics[width=0.9\textwidth]{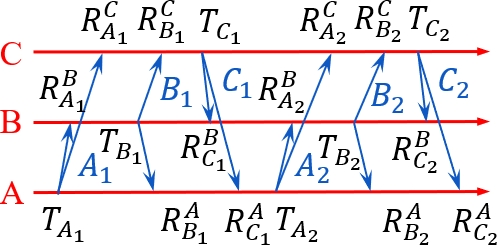}
        \caption{Each side periodically broadcasts \emph{ranging messages}.}
    \end{subfigure}
    \begin{subfigure}[b]{0.23\textwidth}
        \centering
        \includegraphics[width=0.9\textwidth]{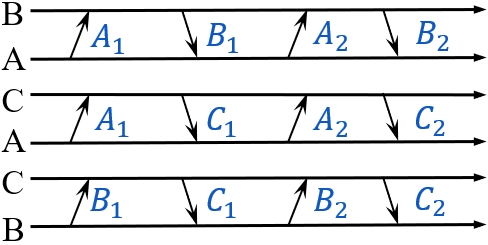}
        \caption{Each pair has enough timestamps to calculate the ToF.}
    \end{subfigure}
       \caption{An illustration of the swarm ranging protocol for the three-side example.}
       \label{fig:toyexample}
\end{figure}

Although this protocol works well, it has not been designed for lightweight aerial intra-swarm ranging in mind. 
As the number of pairs of robots in a swarm explodes in a combinatorial way, the communication bandwidth of lightweight robots saturates easily. 
This results in slow ranging, which in turn severely limits the number of robots in the swarm.
Therefore, the proposed swarm ranging protocol designs a single type of broadcasted message called the \emph{ranging message} for lightweight aerial swarms. 
This broadcasted message is periodically transmitted to the swarm by each participant of the swarm (also called \emph{side}). To illustrate the basic idea of the protocol, we provide a three-side toy example in Fig.~\ref{fig:toyexample}.

\begin{figure}[h]
\centering
\centerline{\includegraphics[width=0.47\textwidth]{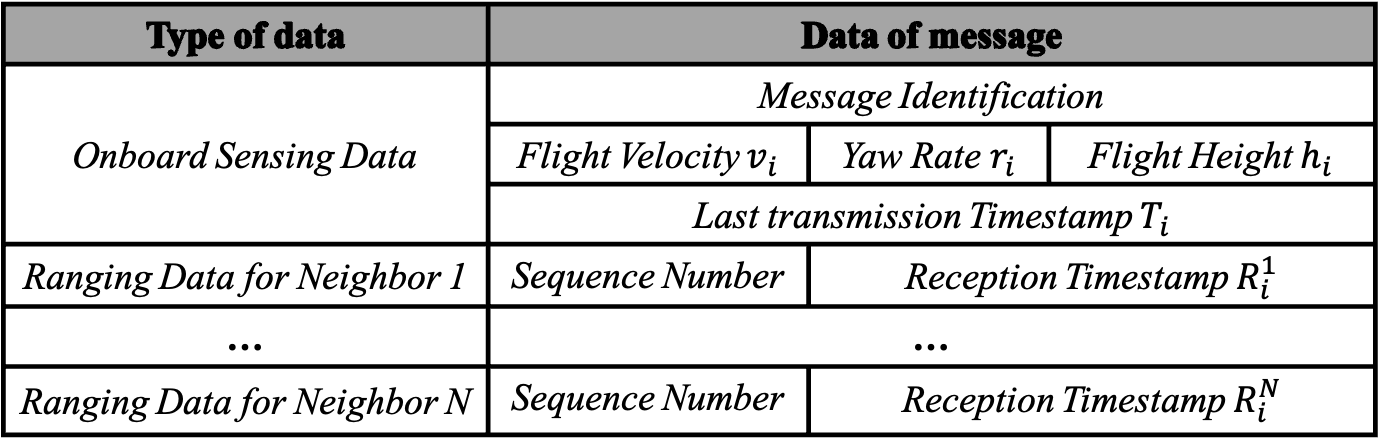}}
\caption{
The proposed ranging message that supports relative localization. 
}
\label{fig:ranging_data_message}
\end{figure}

Three participants A, B, and C take turns to transmit six messages.
Each message can be received by the other two sides because of the broadcast nature of wireless communication.
Then each message generates three timestamps, e.g., three timestamps generated by the ranging message $A_{1}$ sent by $A$ is $T_{A_{1}}$, $R^{B}_{A_{1}}$, $R^{C}_{A_{1}}$, as illustrated in Fig.~\ref{fig:toyexample}(a).
It can be observed that each pair participating has two rounds of message exchanges as in Fig.~\ref{fig:toyexample}(b).
Consequently, there are sufficient timestamps to calculate the ToF for each pair. 

Each ranging message includes the message identification, i.e., sender id and message sequence number, the transmission timestamp of the latest sent ranging message and reception timestamps from all neighbors.

To support relative localization, the ranging message must carry additional velocity, yaw rate, and height information of the sender, so that the neighbors can utilize such information for relative localization estimations.
Therefore, the ranging message data fields are illustrated in Fig.~\ref{fig:ranging_data_message}. 

\begin{figure}[h]
\centering
\centerline{\includegraphics[width=0.47\textwidth]{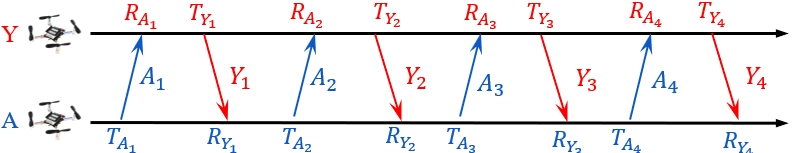}}
\caption{
Message exchanging between $A$ and $Y$ for performing continuous ranging. 
}
\label{fig:slidewindow}
\vspace*{-0.4cm}
\end{figure}
\begin{figure}[h]
\centering
\centerline{\includegraphics[width=0.375\textwidth]{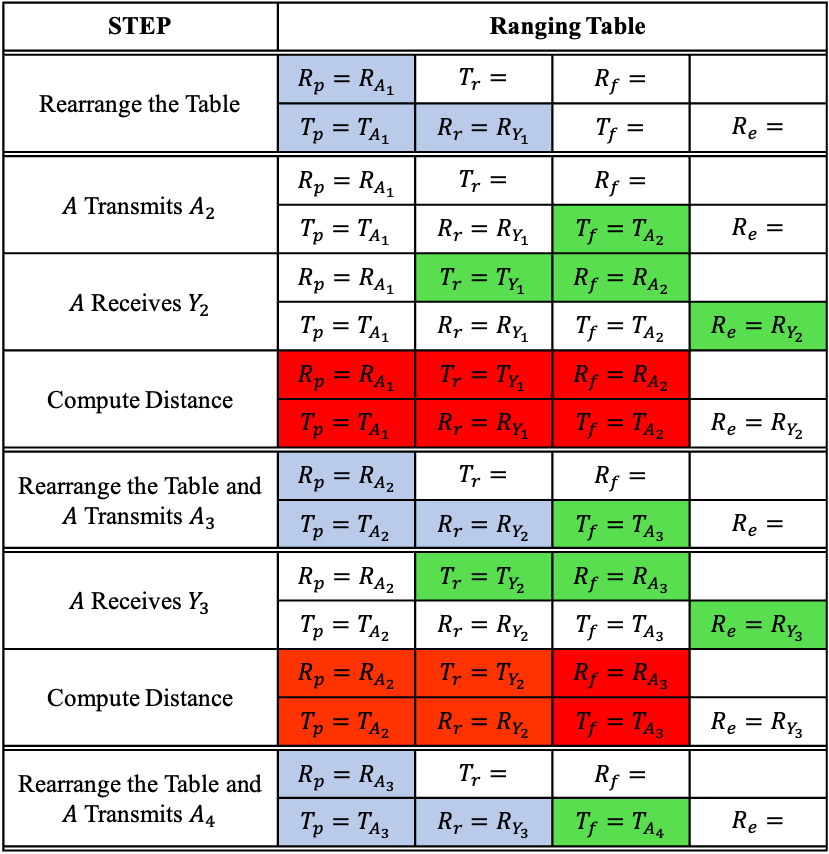}}
\caption{
The swarm ranging steps.
}
\vspace*{-0.4cm}
\label{fig:swarm_ranging_steps}
\end{figure}

As mentioned above, six timestamps ($T_p$, $R_p$, $T_r$, $R_r$, $T_f$, $R_f$) are needed to calculate the ToF. 
Therefore, for each neighbor, an additional data structure is designed to store these timestamps which we named the \textit{ranging table}. 
Assume $A$ is ranging with a number of neighbors, and let $Y$ be any neighbors.
The message exchange between $A$ and $Y$ is depicted in Fig.~\ref{fig:slidewindow}.
Fig.~\ref{fig:swarm_ranging_steps} shows how the ranging messages are generated and the ranging tables are updated to correctly compute the distance between $A$ and $Y$.

\section{Tiny Relative Localization Scheme and Observability}
\label{sec:uwb_filterandobservability}


\subsection{Tiny Relative Localization Scheme}
It is well known that Extended Kalman Filter (EKF) has a very low computational overhead compared to particle filter~\cite{zhang2019improved}, SDP~\cite{li2020robot} and QCQP~\cite{nguyen2023relative}. Therefore, we choose EKF for estimation because it is efficient, which is vital for lightweight-robots with limited computation power. We explain the EKF in detail, as it lays the basis for the ensuing observability and stability analysis. Specifically, we use a discrete model $F(\hat{\vect{X}}_k, \vect{U}_k)$ of \eqref{eq:uwb_model}:
\begin{equation} \label{eq:uwb_EKF_predict}
\begin{aligned}
  \hat{\vect{X}}_{k+1|k}&=F(\hat{\vect{X}}_k, \vect{U}_k)=\hat{\vect{X}}_k+\dot{\vect{X}}_k\Delta{t}, \\
  \vect{P}_{k+1|k}&=\vect{A}_k\vect{P}_{k|k}\vect{A}^T_k+\vect{B}_k\vect{Q}_k\vect{B}_k^T,
\end{aligned}
\end{equation}
where $\Delta{t}$ is the interval time of updating the Kalman filter, the predicted state is represented by $\hat{\vect{X}}_{k+1|k}$, and $\hat{\vect{X}}_{k|k}$ is the estimated state at time step $k$. Furthermore, the first equation in \eqref{eq:uwb_EKF_predict} shows the prediction result using the nonlinear model of \eqref{eq:uwb_model}. The second equation denotes the update of error covariance $\vect{P}$ caused by the prediction step and input noise covariance $\vect{Q}$.
$\vect{A}$ and $\vect{B}$ are the Jacobian matrix for updating $\vect{P}$.

After the prediction update, the Kalman filter fuses the predicted state with the distance observation:
\begin{equation} 
\label{eq:uwb_hFunc}
  z=h(\vect{X}_{ij})=\sqrt{\vect{p}^T_{ij}\vect{p}_{ij}}=\sqrt{x_{ij}^2+y_{ij}^2+(h_j-h_i)^2}.
\end{equation}
Therefore, the Jacobian matrix of observation is
\begin{equation} \label{eq:uwb_hJacob}
  \vect{H}=\partial{h}/{\partial\vect{X}}=[x_{ij}/z, y_{ij}/z, 0].
\end{equation}
The rest of the Kalman filter process is shown as follows.
\begin{equation} \label{eq:uwb_EKFmeasure}
\begin{aligned}
  \vect{K}_k&=\vect{P}_{k|k-1}\vect{H}^T_k(\vect{H}_k\vect{P}_{k|k-1}\vect{H}^T_k+\vect{R}_k)^{-1}\,,\\
  \hat{\vect{X}}_k&=\hat{\vect{X}}_{k|k-1}+\vect{K}_k(z_k-\vect{H}_k\hat{\vect{X}}_{k|k-1})\,,\\
  \vect{P}_k&=(\vect{I}-\vect{K}_k\vect{H}_k)\vect{P}_{k|k-1}.
\end{aligned}
\end{equation}
where $\vect{K}$ is the Kalman gain. Here, both $\vect{Q}$ and $\vect{R}$ are noise covariance parameters and can be formulated as diagonal matrices denoted by $\vect{Q} = \mathrm{diag}([q_v^2, q_v^2, q_r^2, q_v^2, q_v^2, q_r^2])$ and $\vect{R} = \mathrm{diag}([r_d^2])$. $q_v$, $q_r$ and $r_d$ denote the standard deviation of the velocity, yaw rate and distance measurements.

\subsection{Observability Analysis}
\label{sec:uwb_observabilityanalysis}
Since the measurement dimension is lower than the state dimension, the system is only observable under some specific robot behaviors. In order to find out and avoid the nonlinear unobservable robot behaviors for more precise estimation, the Lie derivative method is employed here.

The observability matrix $\vect{O}$ is composed of different orders of Lie derivatives as:
\begin{equation} \label{eq:uwb_obserMatrix}
  \vect{O}=\begin{bmatrix} \nabla{\mathcal{L}_f^0}h \\ \nabla{\mathcal{L}_f^1}h \\ \nabla{\mathcal{L}_f^2}h \end{bmatrix}=
  \begin{bmatrix} (\partial{\mathcal{L}_f^0h})/(\partial{\vect{X}}) \\ (\partial{\mathcal{L}_f^1h})/(\partial{\vect{X}}) \\ (\partial{\mathcal{L}_f^2h})/(\partial{\vect{X}}) \end{bmatrix},
\end{equation}
where $\mathcal{L}_fh$ is the Lie derivative of the model function $f$, and $\nabla{\mathcal{L}_f}h$ is the differential operator of the Lie derivatives.
For simplicity, the power form $\vect{p}_{ij}^T\vect{p}_{ij}/2$ is taken as $\mathcal{L}_f^0h$.
Substituting the system model and observation function, we get:

\begin{equation} \label{eq:uwb_obserDet}
\begin{aligned}
  \lvert \vect{O} \rvert &= -\vect{p}_{ij}^TRS\vect{v}_j(\vect{v}_i^TSr_i+r_j\vect{v}_j^TS^TR^T)S\vect{p}_{ij} \\
            &-(2\vect{v}_i^TRS\vect{v}_j+\vect{p}_{ij}^TR\vect{v}_jr_j)(-\vect{v}_i^T+\vect{v}_j^TR^T)S\vect{p}_{ij}.
\end{aligned}
\end{equation}
which is the determinant of the observability matrix. According to the local weak observability theory, the system is observable only if observability matrix $\vect{O}$ is full rank. In other words, \eqref{eq:uwb_obserDet} should be non-zero.

Although it is difficult to get the full analytical solution of $\lvert \vect{O} \rvert \neq 0$, we can extract three intuitive and practical unobservable conditions. 1) $\vect{p}_{ij}$ is close to zero, which means compact movements will cause lower estimation accuracy;
2) $\vect{v}_j$ is zero, leading that $i^{\mathrm{th}}$ robot cannot find out the heading of $j^{\mathrm{th}}$ robot;
3) The relative velocity $-\vect{v}_i^T+\vect{v}_j^TR^T$ is zero, which happens in most formation flights.
Based on the above resource-constrained method, we will analyze surprising theoretical results in the following subsections.



\subsection{Stochastic Initialization and Stability}
\label{sec:uwb_initialization}
Most swarm robotic studies that use ranging for localization assume known initial relative positions and orientations between the robots \cite{zhou2022swarm}.
However, in practical scenarios the initial relative states $\vect{X}^0_{ij}$ between robots are usually unknown. Manual measurements of the initial positions are time-consuming and make the system less autonomous. Therefore, an automatic initialization method is designed to have the relative localization errors approximate zero before executing cooperative tasks.

For simplicity, we assume the control input of the yaw rate for each robot remains zero during the whole flight, i.e., $r_i=r_j=0$. 
As the drones are in control of their yaw rates, this assumption can be made true by design.
Consequently, we refer to this assumption as the zero yaw rates assumption.

With this assumption, the observability determinant is reduced to
\begin{equation} \label{eq:uwb_obsDetReduce}
\lvert \vect{O}_r \rvert =         -2\vect{v}_i^TRS\vect{v}_j(-\vect{v}_i^T+\vect{v}_j^TR^T)S\vect{p}_{ij}.
\end{equation}

Denote the control inputs for each robot as $\vect{u}_i=[v_i^x, v_i^y, r_i]^T$. The initialization inputs are set to:
\begin{equation} \label{eq:uwb_initInputs} {\medmuskip=0mu
    \vect{u}_i(t) = \begin{cases} 
[v_{\rm{xR}}, v_{\rm{yR}}, 0]^T , & t \in [2kT, (2k+1)T)\\
-[v_{\rm{xR}}, v_{\rm{yR}}, 0]^T, & t \in [(2k+1)T, 2(k+1)T)
\end{cases} }
\end{equation}
where $v_{\rm{xR}}$ and $v_{\rm{yR}}$ are two-axis velocities generated randomly within the range of $(0, v_{\rm{max}}]$ at time $t=2kT$ in local clocks.
This initialization process prevents the robots from flying away from the initial position in a short time such that a safe flight is guaranteed.

We will first study what the state estimate converges to in observable conditions and then in unobservable conditions.

\begin{theorem} \label{thm:uwb_initObs}
If the system input satisfies $R(\psi_{ij})\vect{v}_j-\vect{v}_i \neq 0$, all relative states of the Kalman filter converge and are exponentially bounded.
\end{theorem}

\begin{proof}
The observability determinant \eqref{eq:uwb_obsDetReduce} is not zero when $R(\psi_{ij})\vect{v}_j-\vect{v}_i \neq 0$.
Therefore, the system satisfies the nonlinear observability rank condition.
According to \cite{reif1999stochastic}, the corresponding estimator converges exponentially and the estimation error is bounded.
The detailed convergence proof is omitted for the weak observable systems as many references have already proved it.
\end{proof}

\begin{theorem}\label{thm:uwb_init_unObs}
For multiple robots with dynamic estimation model \eqref{eq:uwb_model}, if the control inputs follow the initialization process \eqref{eq:uwb_initInputs} and there is an unobservable condition $R(\psi_{ij})\vect{v}_j-\vect{v}_i=0$, then the estimated relative state of the Kalman filter will converge to an unobservable subspace, i.e.
\begin{equation} \label{eq:uwb_thmInitUnobs}
     \lim_{t\to\infty}\hat{\vect{X}}_{ij}(t) \to
     \{x, y, \psi|\sqrt{x^2+y^2}=z_{\mathrm{GT}}, \psi=\psi_{\mathrm{GT}}\},
\end{equation}
and all states $[x_{ij}, y_{ij}, \psi_{ij}]^T$ drift slowly once they reach the subspace. $z_{\mathrm{GT}}$ and $\psi_{\mathrm{GT}}$ denote a constant distance measurement and constant relative yaw.
\end{theorem}

\begin{proof}
The derivative of the estimate state $\hat{\vect{X}}$ can be written as:
\begin{equation} \label{eq:uwb_derHatX}
    \dot{\hat{\vect{X}}}_{ij}=f(\hat{\vect{X}}_{ij}, \vect{U}_{ij})+\vect{K}(z-h(\hat{\vect{X}}_{ij})).
\end{equation}
According to \cite{reif1998ekf}, the Kalman gain $\vect{K}$ and the derivative of the error covariance matrix $\vect{P}$ can be represented by
\begin{equation} \label{eq:uwb_KfromRef}
\begin{aligned}
    \vect{K}&=\vect{P}\vect{H}^T\vect{R}^{-1}\\
    \dot{\vect{P}}&=\vect{A}\vect{P}+\vect{P}\vect{A}^T-\vect{P}\vect{H}^T\vect{R}^{-1}\vect{H}\vect{P}+\vect{B}\vect{Q}\vect{B}^T.
\end{aligned}
\end{equation}
Based on the definition of the Kalman function, the optimal gain $\vect{K}$ always satisfies the following equations:
\begin{equation} \label{eq:uwb_Kdefinition}
    \frac{\partial{\rm{tr}(\vect{P})}}{\partial{\vect{K}}}=0\text{,}\
    \vect{P}=\rm{cov}(\vect{X}-\hat{\vect{X}})=\rm{cov}(\tilde{\vect{X}})\text{,}
\end{equation}
Therefore, if a \textbf{unique equilibrium} space of state error $\tilde{\vect{X}}$ can be found, the relative estimation under the unobservable condition will converge to that space.

The equilibrium space can be found by setting $\dot{\tilde{\vect{X}}}=\dot{\vect{X}}_{ij}-\dot{\hat{\vect{X}}}_{ij}$ to zero.
$\dot{\vect{X}}_{ij}=[0, 0]^{T}$ can be derived by combining \eqref{eq:uwb_model}, zero yaw rates assumption, and $R(\psi_{ij})\vect{v}_j-\vect{v}_i=0$.
Hence, substitute \eqref{eq:uwb_derHatX} into $\dot{\tilde{\vect{X}}}=-\dot{\hat{\vect{X}}}_{ij}=0$ which yields
\begin{equation} \label{eq:uwb_equili}
\begin{bmatrix}R(\hat{\psi}_{ij})\vect{v}_j-\vect{v}_i\\0\end{bmatrix}+\vect{K}(z-h(\hat{\vect{X}}))=0.
\end{equation}
A two-dimensional time-invariant solution for Eq. \ref{eq:uwb_equili} is:
\begin{equation} \label{eq:uwb_equiliSpace}
\begin{cases} 
\hat{x}_{ij}^2+\hat{y}_{ij}^2=z^2_{\rm{GT}}\text{,}\\ \hat{\psi}_{ij}=\psi_{\rm{GT}}.
\end{cases}
\end{equation}
Here we prove that \eqref{eq:uwb_equiliSpace} is the unique time-invariant solution by studying all cases.
Case 1: $R(\hat{\psi}_{ij})\vect{v}_j-\vect{v}_i=0$ and $\vect{K}=0$; Case 2: $R(\hat{\psi}_{ij})\vect{v}_j-\vect{v}_i=0$ and $z-h(\hat{\vect{X}})=0$; Case 3: $R(\hat{\psi}_{ij})\vect{v}_j-\vect{v}_i\neq 0$ and $\vect{K}(z-h(\hat{\vect{X}}))\neq 0$, but they sum to zero.

Case 1 holds only if $\vect{P}\vect{H}^T=0$ according to \eqref{eq:uwb_KfromRef}, which furthermore leads to $\dot{\vect{P}}=\vect{A}\vect{P}+\vect{P}\vect{A}^T+\vect{B}\vect{Q}\vect{B}^T$. Hence, $\vect{P}$ is independent of distance measurement $z$ from \eqref{eq:uwb_EKF_predict}, while $\vect{H}$ is dependent on $z$ from \eqref{eq:uwb_hJacob}. Since $\vect{H}$ always varies over time due to measurement noise while $\vect{P}$ does not, $\vect{P}\vect{H}^T=0$ will be a transient condition.
Case 2 corresponds to the time-invariant solution in \eqref{eq:uwb_equiliSpace}.
In case 3, $\vect{K}$ is time variant as it contains the integration of state variables which are in matrix $\vect{A}$, $\vect{B}$, and $\vect{H}$ according to \eqref{eq:uwb_KfromRef}. Thus, this solution is also transient.
Therefore, \eqref{eq:uwb_equiliSpace} is the unique time-invariant equilibrium state space, and the estimated states will converge to the equilibrium space as shown in \eqref{eq:uwb_thmInitUnobs}.
\end{proof}

The initialization procedure works mostly under observable conditions because of the velocity inputs that are picked randomly and started asynchronously. This makes the probability that two robots are flying with continuously identical velocities at the same times extremely unlikely. Hence, the initialization procedure ensures with high probability that the state $\hat{\vect{X}}$ converges to the true values after a finite-time flight.

\subsection{Self-regulated Estimation Convergence}
\label{sec:uwb_self_converge}

In a formation flight, the reference setpoint is $\bar{\vect{p}}_{ji}$ for the $i^{\rm{th}}$ robot in the frame of the $j^{\rm{th}}$ robot.
Thus, the control error of the relative position is
\begin{equation} \label{eq:uwb_ctrlErr}
    \vect{e}_{ij}=\hat{\vect{p}}_{ij}-\bar{\vect{p}}_{ij}=\hat{\vect{p}}_{ij}+R(\hat{\psi}_{ij})\bar{\vect{p}}_{ji},
\end{equation}
where $\hat{\vect{p}}_{ij}$ is the relative position estimation. Considering the relative system dynamics
$\dot{\vect{p}}_{ij}=R(\psi_{ij})\vect{v}_j-\vect{v}_i-Sr_i\vect{p}_{ij}$, a dynamic inversion formation control law is proposed as
\begin{equation} \label{eq:uwb_ndi_ndi}
    \vect{v}_i=k_{\mathrm{c}}\vect{e}_{ij}+R(\hat{\psi}_{ij})\vect{v}_j-Sr_i\hat{\vect{p}}_{ij},
\end{equation}
where $k_{\mathrm{c}}$ denotes the control gain, which leads to $\lim \vect{e}_{ij}\to0$. Therefore, the real relative position $\vect{p}_{ij} \approx \hat{\vect{p}}_{ij}=-\vect{e}_{ij}-R(\psi_{\rm{GT}})\bar{\vect{p}}_{ji}$ approximates a constant, and the following holds:
\begin{equation} \label{eq:uwb_formUnobs}
    \dot{\vect{p}}_{ij}=0=R(\psi_{ij})\vect{v}_j-\vect{v}_i.
\end{equation}
This leads to a zero determinant in \eqref{eq:uwb_obsDetReduce} such that the stable states of formation control cause an unobservable condition for the relative estimation system.

\begin{problem} \label{pb:uwb_drift}
The input and measurement noise is omnidirectional and hence typically has a component tangent to the unobservable circle trajectory.
This leads to the estimation drift of relative states, hence $\hat{\vect{p}}_{ij}\neq \bar{\vect{p}}_{ij}$.
\end{problem}

\begin{theorem} \label{thm:uwb_formUnobsProof}
Given the converged state estimation $\vect{p}_{ij}$ and $\psi_{ij}$, according to Theorem \ref{thm:uwb_init_unObs}, the invariant $\hat{\psi}_{ij}$ and the estimation drift in Problem \ref{pb:uwb_drift}.
The estimation error will remain converged and bounded even if the multi-robot system is under unobservable maneuvers such as the formation flight.
\end{theorem}

\begin{proof} \label{pr:uwb_selfRegulated}
After the initialization and the formation control, relative states satisfy $\vect{p}_{ij} = \hat{\vect{p}}_{ij} = \bar{\vect{p}}_{ij}$. There are two unobservable cases.

\textbf{Case 1:} Define the estimation drift in Problem \ref{pb:uwb_drift} as $\Delta \vect{p}_{ij}$. The incorrect relative estimation has the following relationship to the real and reference relative positions:
\begin{equation} \label{eq:uwb_self_estiNeqRef}
    \hat{\vect{p}}_{ij} = \Delta \vect{p}_{ij} + \vect{p}_{ij} \neq \bar{\vect{p}}_{ij}.
\end{equation}
Substitute \eqref{eq:uwb_ctrlErr} into \eqref{eq:uwb_ndi_ndi}, and consider the zero yaw rate assumption, we can get
\begin{equation} \label{eq:uwb_self_vIsObs}
    \vect{v}_i=k_{\mathrm{c}}(\hat{\vect{p}}_{ij}-\bar{\vect{p}}_{ij})+R(\hat{\psi}_{ij})\vect{v}_j.
\end{equation}
In view of \eqref{eq:uwb_formUnobs} and \eqref{eq:uwb_self_estiNeqRef}, the state will become observable again due to the ensuing control actions: 
\begin{equation} \label{eq:uwb_self_vNeq0}
    R(\psi_{ij})\vect{v}_j-\vect{v}_i = k_{\mathrm{c}}(\bar{\vect{p}}_{ij}-\hat{\vect{p}}_{ij}) \neq 0.
\end{equation}
Hence, based on Theorem \ref{thm:uwb_initObs}, the estimated relative position $\hat{\vect{p}}$ will converge again to the real value $\vect{p}$.

\textbf{Case 2:} The system is possibly unobservable when $\hat{\vect{p}}=\bar{\vect{p}}$ but $\hat{\vect{p}} \neq \vect{p}$, which means the relative estimation is incorrect and the system is unobservable. In this case, the relative position will converge to the subspace (the circle trajectory) according to Theorem \ref{thm:uwb_init_unObs}. However, the  measurement noise on $\vect{v}_1$ and $\vect{v}_i$ is omnidirectional, so it has components orthogonal to the equilibrium state, leading to case 1 and hence observability. Moreover, external disturbances and actuation noise will lead to non-zero $R(\psi_{i1})\vect{v}_1-\vect{v}_i$, and hence observability.
\end{proof}

Our paper presents the surprising theoretical stability proof for observability in EKF-based relative localization.
This proof, grounded in the mathematical formulation of the EKF, is crucial for understanding the convergence behavior in aerial swarms.

\section{Simulated Relative Localization}
\label{sec:uwb_sim}
This section shows simulation results for the relative localization including efficacy, accuracy, and convergence stability under unobservable conditions.
\subsection{Localiation Performance}

In this simulation, the relative state between two robots is estimated and compared to the ground-truth relative position and yaw to verify the localization accuracy.

\begin{figure}[ht]
    \centering
    \includegraphics[width=0.48\textwidth, trim={0.8cm 0cm 1cm 1cm}, clip]{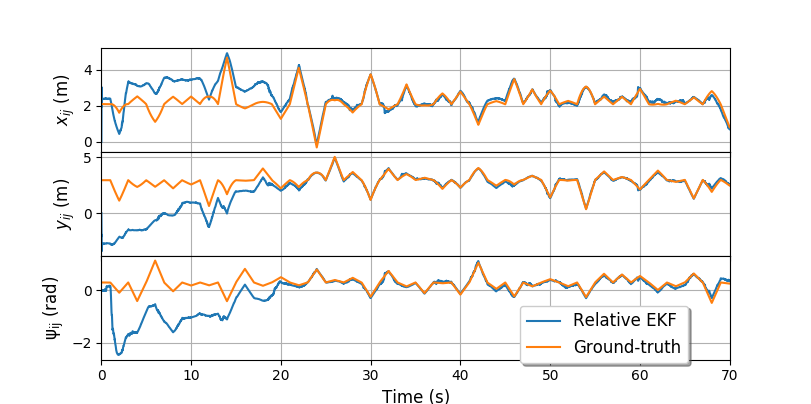}
    \caption{Simulation results of the relative state estimation between two robots on $x_{ij}, y_{ij}$, and $\psi_{ij}$. Both robots are randomly initialized at unknown position and yaw, and they are 2 meters far away each other. Then each robot flies a start-up procedure with 2-second periodic random settings of velocity and yaw rate (as illustrated in Section \ref{sec:uwb_initialization}).}
    \vspace*{-0.1cm}
    \label{fig:uwb_simXYYaw}
\end{figure}

This simulation is configured with a time interval $dt=0.01s$ and a maximum moving velocity of 1 m/s.
The settings of the simulation are: input noise deviation of 0.25 m/s and 0.01 rad/s, and a distance measurement deviation 0.1 m.
The initial estimated states are set to zero, while the ground-truth initial states are set randomly and uniformly in a range of [-3,3] m and [-1,1] rad.
The parameters of the relative EKF are set to be $\vect{Q}=\mathrm{diag}([0.25^2, 0.25^2, 0.4^2, 0.25^2, 0.25^2, 0.4^2])$, $\vect{R}=0.1^2$, and $\vect{P}=\mathrm{diag}([10, 10, 0.1])$, based on the simulated estimation performance.

\begin{figure}[ht]
    \centering
    \includegraphics[width=0.48\textwidth, trim={1cm 0cm 1cm 1cm}, clip]{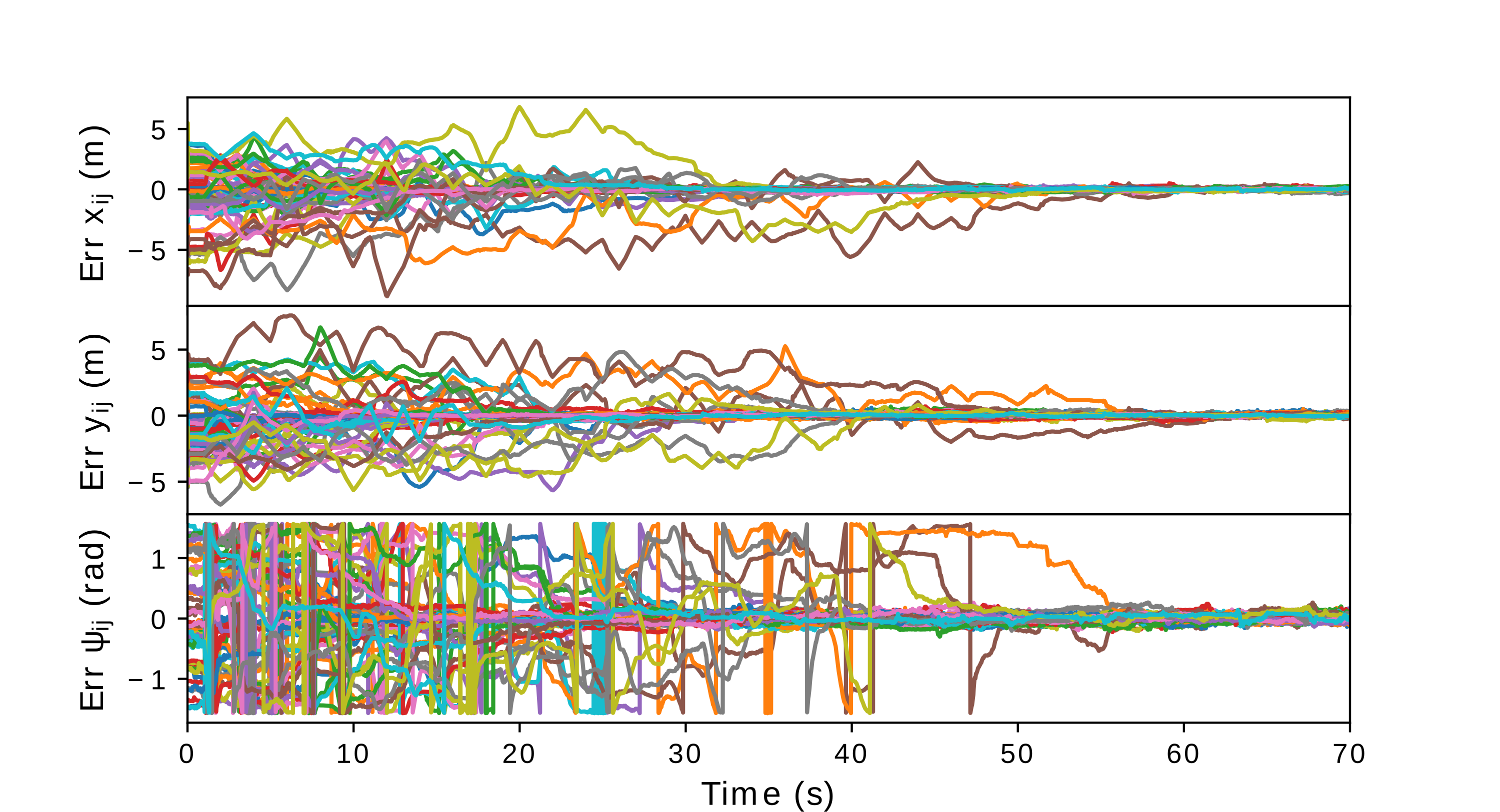}
    \caption{Simulation results of estimation error convergence. 3 dimensional relative states are shown from 50 tests with different configurations. Each line with different color represents a different estimation test in three states of $x_{ij}$, $y_{ij}$, and $\psi_{ij}$. All errors are calculated by comparing the estimated states with the ground-truth.}
    \label{fig:uwb_simErr50}
\end{figure}

Relative localization results are shown in Fig. \ref{fig:uwb_simXYYaw}, where we can see that the relative position and yaw approximate the ground-truth after a random flight.
The orange line represents the real relative states, while the blue line means the relative states estimated by EKF.
With 50 different random tests as shown in Fig. \ref{fig:uwb_simErr50}, the average convergence time is 20 seconds.
Overall, the average convergence time is short, and the localization is accurate for multi-robot control.

\subsection{Unobservability and Self-regulated Convergence}
We study the influence of unobservable flight behavior on the relative localization \emph{after} estimation convergence in practice.
Two situations that lead to unobservability will be discussed: 1) Formation flight that causes $-\vect{v}_i^T+\vect{v}_j^TR^T=0$, while the yaw rates of $r_i$ and $r_j$ remain zero; 2) the $j^{\mathrm{th}}$ robot has zero velocity, i.e. $\vect{v}_j=0$, so that the relative yaw $\psi_{ij}$ should be unobservable.

\begin{figure}[ht]
    \centering
    \includegraphics[width=0.48\textwidth, trim={0cm 1.8cm 0cm 0.8cm}, clip]{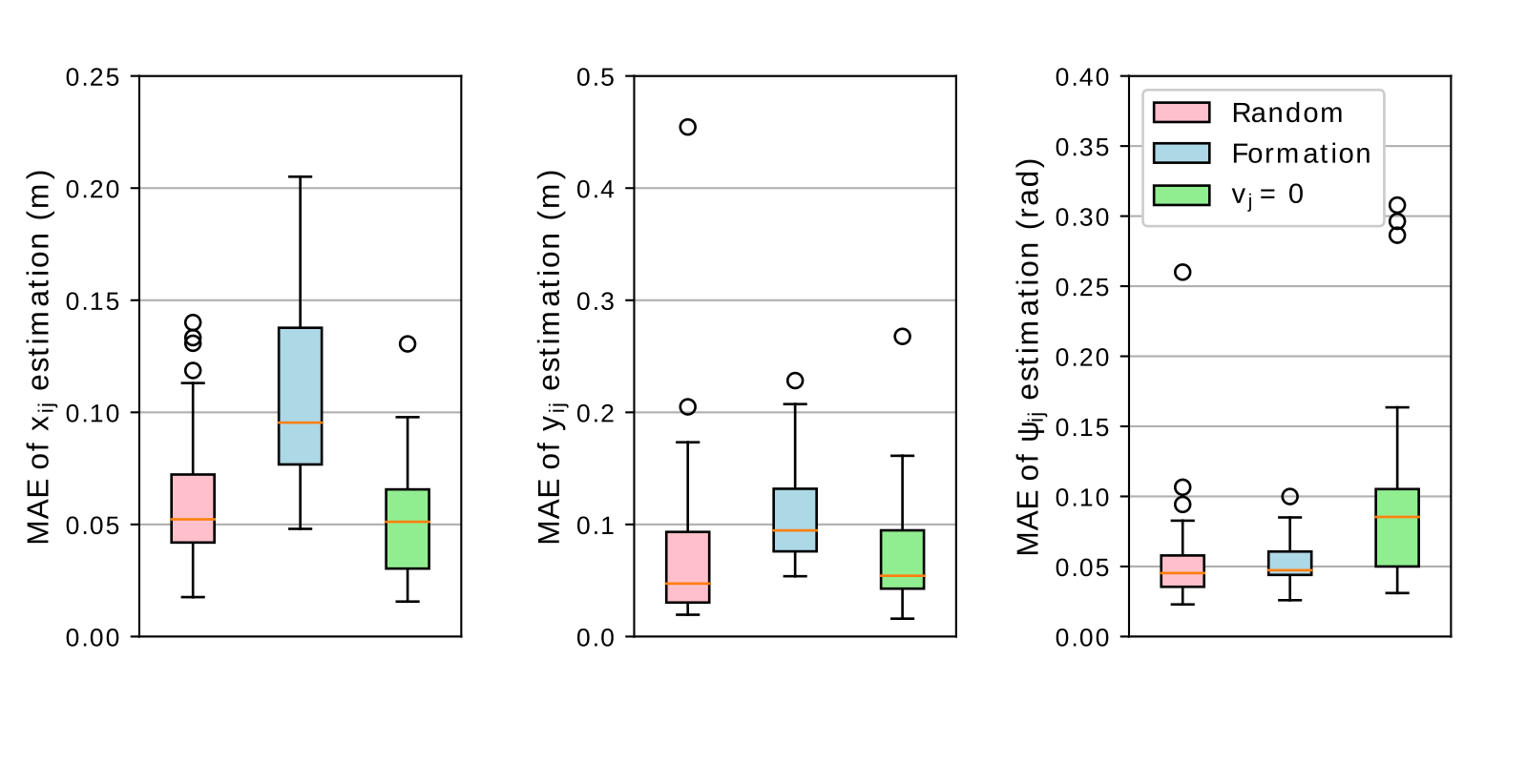}
    \caption{Error distribution of relative localization in different unobservable situations. For each situation, the mean absolute errors are obtained from 50 different tests, during the 20 seconds after the estimation convergence. Boxes of red, blue, and green color represent random flight, formation flight, and zero velocity of the $j^{\mathrm{th}}$ robot, respectively.}
    \vspace*{-0.1cm}
    \label{fig:uwb_simObserve}
\end{figure}

In Fig. \ref{fig:uwb_simObserve}, relative localization performance with unobservability-inducing control inputs are shown. 
By comparing the red and blue boxes, the effects of formation flight are: 1) An increase of estimation errors on all relative states; 2) The relative estimation is still rather accurate with position error less than 0.2 m. This validates the self-regulated estimation convergence theory in Section \ref{sec:uwb_self_converge}.
The result indicates that once the estimation is not correct, robots will deviate from their role in the formation (in terms of velocity and position), which in turn makes the system observable again.

In the case that the robot to be localized has zero velocity, the relative yaw estimation has a larger error compared to normal random flight, which can be seen from the green box.
However, as indicated in Section \ref{sec:uwb_observabilityanalysis}, the relative position of $x_{ij}$ and $y_{ij}$ is still observable.
Therefore, the green boxes show an estimation error similar to the red boxes in axes of $x_{ij}$ and $y_{ij}$.

\section{Real-world Experimental Results}
\label{sec:uwb_exp}
This section presents the real-world performance for our proposed ranging protocol and relative localization method.

The swarm of the aerial robot system consists of commercial Crazyflie 2.1 quadrotor.
Each quadrotor is equipped with a 3-axis accelerometer, 3-axis gyroscope, flow deck (VL53L1x height sensor and PMW3901 optical flow sensor), and loco deck (DWM3000 ultra-wideband sensor).
The flow sensor can provide velocity at 100 Hz, and the distance measurement frequency can reach over 16 Hz.
The processor is an STM32F4 running at 168 MHz with 192 KB memory, on which both relative estimation, control and swarm ranging are running.
The OptiTrack motion capture system is used for tracking the ground-truth position and yaw of each robot.
The ground-truth data is only used for post-processing to validate the relative estimation performance.


\subsection{Communication and Ranging Performance }
\begin{figure}[ht]
    \centering
    \begin{subfigure}[b]{0.3\textwidth}
        \centering
        \includegraphics[width=\textwidth, trim={0cm 0.5cm 0cm 0cm}]{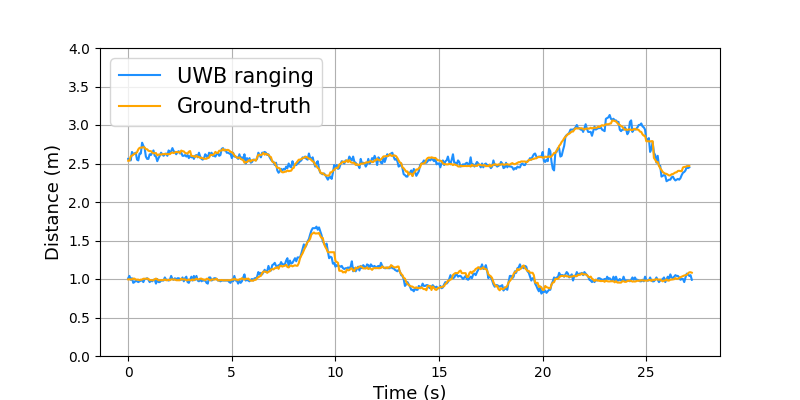}
        \caption{UWB measurements}
        \label{fig:uwb_distance}
    \end{subfigure}
    \hfill
    \begin{subfigure}[b]{0.15\textwidth}
        \centering
        \includegraphics[width=\textwidth, trim={1cm 0.5cm 0cm 0.2cm}]{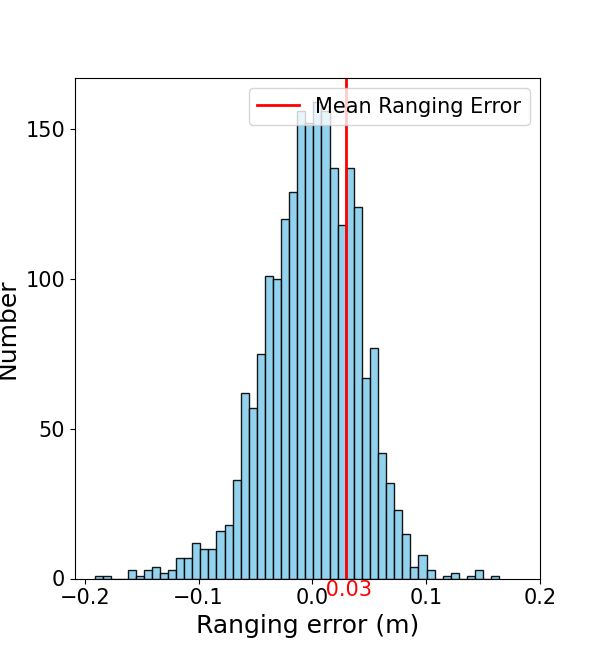}
        \caption{Error distribution}
        \label{fig:uwb_distance_error}
    \end{subfigure}
       \caption{These measurements come from a swarm of 3 Crazyflie quadrotors with a 25-seconds flight. (a) presents the outlier-rejected UWB swarm ranging data depicted by the blue line, alongside the ground-truth distance from Lighthouse represented by the orange line. (b) shows the error distribution in distance measurements against the Lighthouse's ground-truth, with the red line marking the mean absolute error is about 0.03 meters.}
       \label{fig:uwb_distance_measure}
       \vspace*{-0.35cm}
\end{figure}

As illustrated in Fig. \ref{fig:uwb_distance_measure}, 
A median filter is applied to reject the outliers of the raw ranging measurements from UWB. As shown in Fig. \ref{fig:uwb_distance_measure}, the distance measurement mean absolute error between two robots is
about 0.03 m. 
This ranging technique has more accurate and less-biased measurements than that in \cite{van2019board}.

In Table.~\ref{tab:diffRecep}, even with more than 20 drones, the packet reception rate can still be maintained at around 80 percent.
Moreover, according to Table.~\ref{tab:RX_processing}, the time delay required to process a ranging message increases as the neighbors number increases. This is due to the ranging message contains information for each neighbor. And the value is around 0.5 ms if there are 26 participants for relative localization, which clearly shows its scalability.
The above data clearly shows that the proposed ranging protocol is capable to scale to a large swarm.

\begin{figure}[h]
    \centering
    \includegraphics[width=0.30\textwidth]{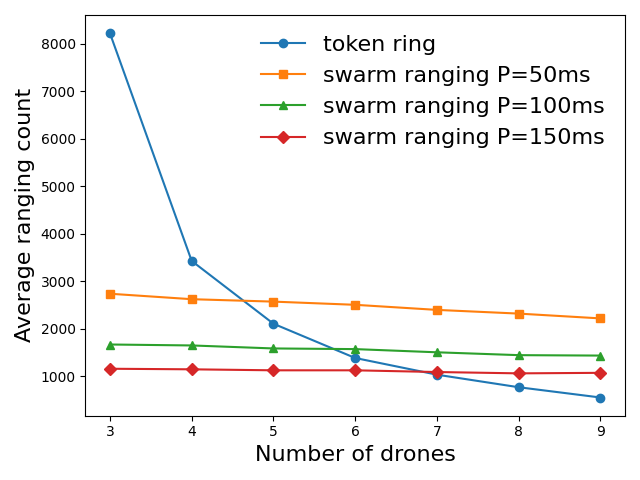}
    \caption{Comparison with ranging based on token ring.}
    \vspace*{-0.1cm}
    \label{fig:RangingComparison}
\end{figure}

\begin{minipage}{0.48\textwidth}
  \captionof{table}{Packet reception rate for different numbers of drones, under two periods: 40ms and 60ms.}
  \label{tab:diffRecep}
  \renewcommand{\arraystretch}{2.5} 
 \resizebox{\linewidth}{!}{
    \begin{tabular}{|c||c|c|c|c|c|c|c|c|c|c|c|c|} 
      \hline
      \large Number of drones  & \large $2$ &  \large $6$ & \large $10$ & \large $14$ & \large $18$& \large $20$& \large $22$& \large $24$\\
      \hline
      \large 40ms Period  & \large $98.99\%$ &  \large $96.59\%$ &  \large $90.74\%$ &  \large $87.75\%$ &  \large $83.84\%$ & \large $81.24\%$ & \large $78.69\%$& \large $77.48\%$\\
      \hline
      \large 60ms Period & \large $99.33\%$ &  \large $98.69\%$ &  \large $96.22\%$ &  \large $92.97\%$  & \large $89.56\%$ & \large $88.66\%$ & \large $86.02\%$ & \large $85.49\%$\\
      \hline
    \end{tabular}

}
\end{minipage}
We compare our swarm ranging protocol to the token ring based ranging algorithm.
We vary the average ranging period $\Bar{P}$, from $50ms$ to $150ms$.
We vary the numbers of drones, from 3 to 9. 
Fig.~\ref{fig:RangingComparison} shows the ranging counts recorded by drone A for $200s$.
Any point in Fig.~\ref{fig:RangingComparison} is the average successful ranging count by drone A with its neighbors.
We can see that the performance of token ring algorithm decreases dramatically with the growth of participants, because it executes sequentially.
While the performance of swarm ranging decreases slightly because the probability of message collision increases as the number increases.
When there are more than 5 drones, our swarm ranging protocol outperforms the token ring algorithm.
When 9 drones participate, the average number for a drone to successfully range with another drone by our protocol is about 5 times higher than that by the token ring algorithm.

According to our experiments, the ranging message transmission period is set to 60 ms.
In other words, we ranging at 16 Hz for 13 robots.
Compared with the state of the art, our work has a higher communication frequency in a swarm with a large number of robots, than 16 Hz and 3 robots in \cite{van2019board}, 40 Hz and 3 robots in \cite{guo2019ultra}, and 10 Hz and 2 robots in \cite{guler2020peer}.
The faster ranging measurement provided by the swarm ranging protocol improves the scalability of the wireless-ranging based relative localization.

\begin{minipage}{0.48\textwidth}
  \captionof{table}{Our measurements of the processing delay to handle a ranging message.}
  \label{tab:RX_processing}
  \renewcommand{\arraystretch}{1} 
 \resizebox{\linewidth}{!}{
    \begin{tabular}{|c||c|c|c|c|} 
      \hline
      \large Number of drones  & \large $2$ & \large $10$ & \large $18$& \large $26$\\
      \hline
      \large RX processing Time (ms) & \large 0.339 & \large 0.441 & \large 0.548 & \large 0.662 \\
      \hline
    \end{tabular}

}
\end{minipage}

\subsection{Relative Localization Performance}

\begin{figure}[ht]
    \centering
    \includegraphics[width=0.48\textwidth, trim={1cm 0cm 1cm 0cm}, clip]{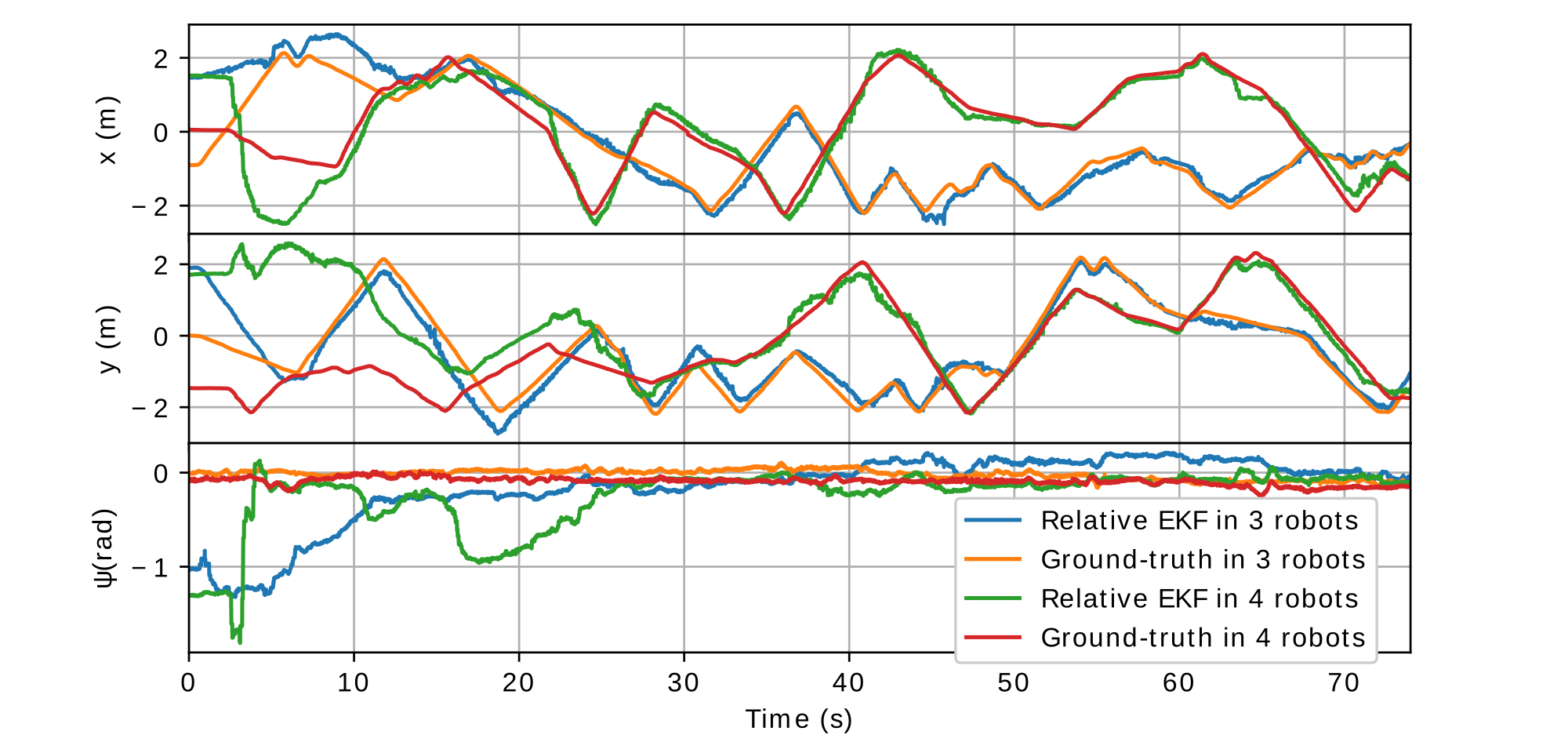}
    \vspace*{-0.2cm}
    \caption{Real-world relative localization in 3-robot and 4-robot systems respectively. Here, $x$, $y$ and $\psi$ denote the absolute XY position and yaw of the $2^{\mathrm{nd}}$ robot, calculated by the relative EKF from the $1^{\mathrm{st}}$ robot, and compared with ground-truth from the OptiTrack motion capture system.}
    \label{fig:uwb_expXYYaw}
\end{figure}

In Fig. \ref{fig:uwb_expXYYaw}, we can see that the unknown initial states can be estimated in about 20 seconds.
After convergence, the estimation errors remain converged and the positioning accuracy is high.
\begin{table}[ht]
\footnotesize
\begin{center}
\caption{Comparisons with the state-of-the-art UWB relative localization methods, including dependencies, accuracy and numbers.}
\label{tab:sensor_difference}
\begin{tabular}{p{0.07\textwidth}>{\centering}p{0.03\textwidth}>{\centering}p{0.04\textwidth}>{\centering}p{0.04\textwidth}>{\centering}p{0.03\textwidth}>{\centering\arraybackslash}p{0.13\textwidth}}
\hline
\multirow{2}{*}{Method}&\multicolumn{4}{c}{Dependence}&\multirow{2}{*}{Accuracy and Number}\\\cline{2-5}
&IMU&Velocity&Height&North&\\
\hline
2019,\cite{guo2019ultra} & $\bullet$ & $\bullet$ & $\bullet$ & $\bullet$ & 40Hz, $\sim$0.2m, 3\\
2020, \cite{van2019board} & $\bullet$ & $\bullet$ & $\bullet$ & & 16Hz, $\sim$0.22m, 3 \\
2020, \cite{guler2020peer} & $\bullet$ & $\bullet$ & $\bullet$ & $\bullet$ & 10Hz, $\sim$1.4m, 2\\
2021, \cite{cossette2021relative} & $\bullet$ & & & $\bullet$ & 10Hz, $\sim$0.6m, 2\\
\textbf{Ours} & $\bullet$ & $\bullet$ & $\bullet$ & & \textbf{16Hz}, $\sim$\textbf{0.13m}, 13\\
\hline
\end{tabular}
\end{center}
\end{table}

Quantitative analysis of the relative localization performance compared to the state-of-the-art technologies is given by Table. \ref{tab:sensor_difference}. These experimental statistic results indicate that our localization system has much higher ranging communication frequency and less localization errors in larger swarms compared to the methods in the literature, and our method does not rely on the north measurement which is unstable in indoor environments.

\begin{figure}[ht]
    \centering
    \includegraphics[width=0.45\textwidth, trim={0cm 0cm 0cm 0cm}, clip]{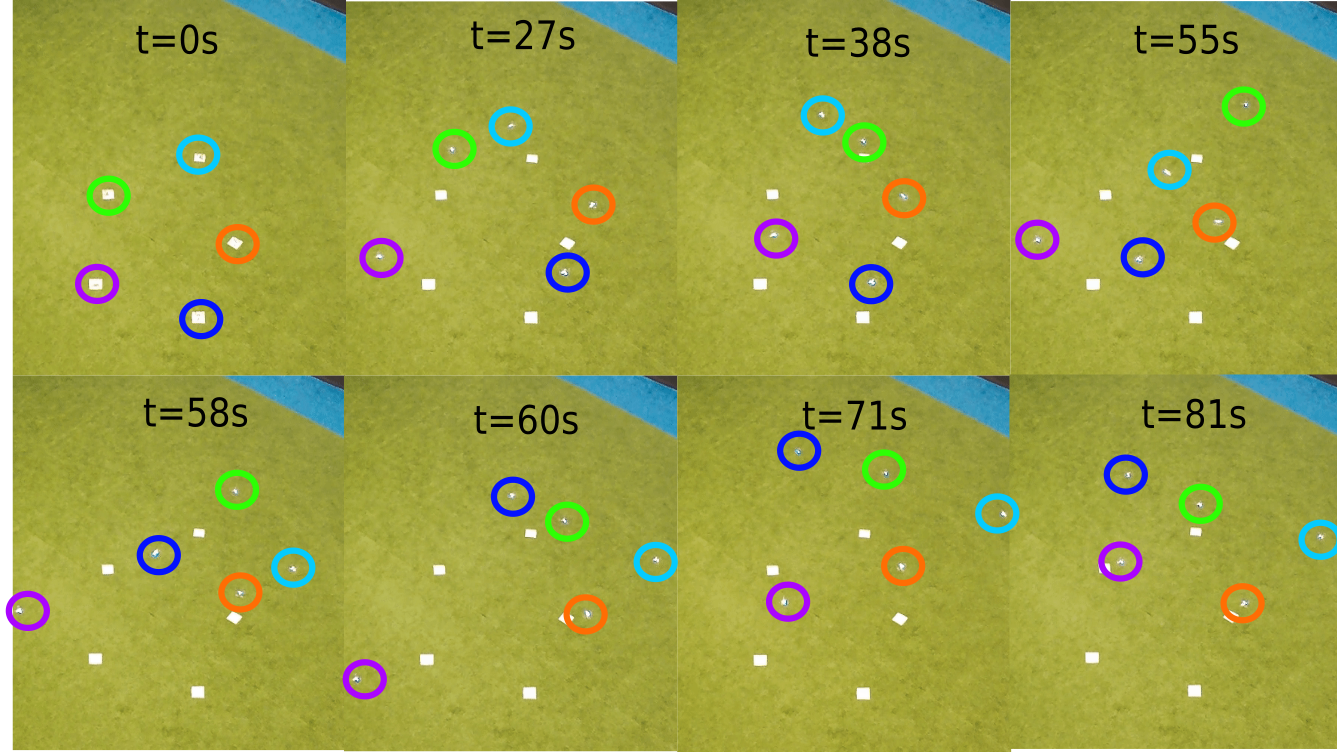}
    \caption{Top view of the formation flight of 5 robots. Eight figures show different flight status such as take-off, initialization procedure, distributed control for formation flight, and hovering. Five circles with different colors show the positions of five tiny drones, respectively. Full flight details can be found in the video link \url{https://www.youtube.com/playlist?list=PL_KSX9GOn2P9sgaX3DHnPsnBCJ76fLNJ5}}
    \label{fig:uwb_expFormation}
\end{figure}

Fig. \ref{fig:uwb_expFormation} shows how a 5 robots team achieves a formation flight based on the proposed initialization method, relative localization and distributed control.
At $t=0$ s, all tiny flying robots take off from 5 random unknown positions with unknown random yaw angles.
After the 30-seconds initialization procedure, all robots start flying to the desired formation positions with respect to the $1^{\mathrm{st}}$ robot with the orange circle.
Starting from $t=60$ s all robots do a formation flight with constant relative positions to the $1^{\mathrm{st}}$ robot which performs a random flight. Finally, five robots form an Olympic-flag-like shape. This shape is maintained by all robots even in the hovering state, which is unobservable for the multi-robot system. 

\begin{figure}[ht]
    \centering
    \includegraphics[width=0.49\textwidth, trim={0cm 0cm 0cm 0cm}, clip]{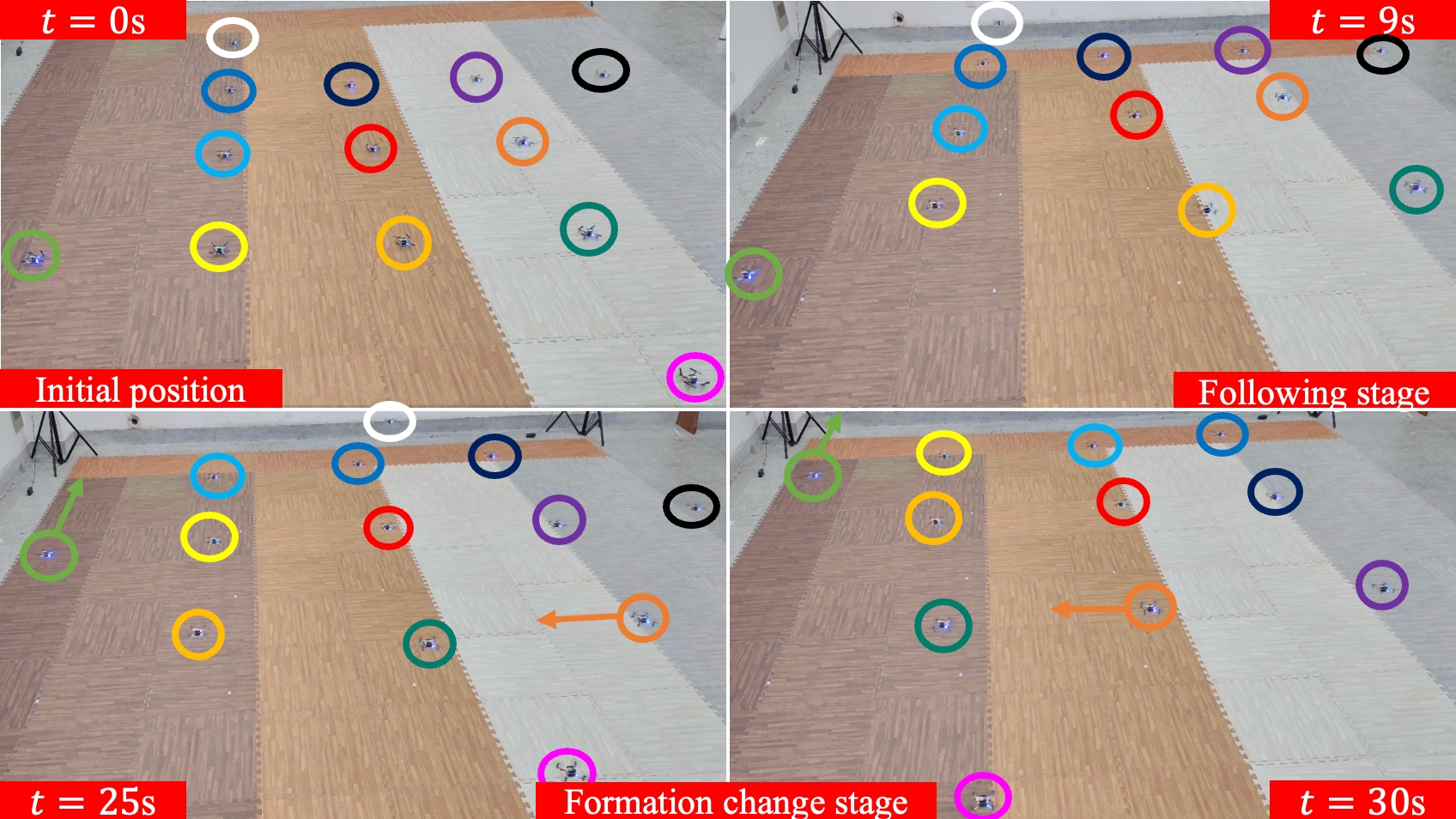}
    \caption{Top view of the formation flight of 13 robots. Four figures show different flight status such as initial position, following stage and Formation change stage. The red circle represents the leader drone and circles of other colors represent follower drones. Full flight details can be found in the video link \url{https://www.youtube.com/playlist?list=PL_KSX9GOn2P9sgaX3DHnPsnBCJ76fLNJ5}}
    \label{fig:uwb_13Formation}
\end{figure}

As shown in Fig. \ref{fig:uwb_13Formation}, we conducted 13 large-scale swarm robots flight experiments. 
Their initial positions are 1 meters apart from each other.
At $t=0$ s, all robots take off from their respective known initial positions.
After a 5-seconds initialization procedure, all drones are flying in formation, maintaining a constant relative position to the $1^{\mathrm{st}}$ robot in the red circle which is flying randomly.
Starting from $t=15$ s, all drones start to change formation, and the arrow drawn in the figure becomes the direction of the next formation change.

\begin{figure}[ht]
    \centering
    \includegraphics[width=0.48\textwidth, trim={0cm 0cm 0cm 0cm}, clip]{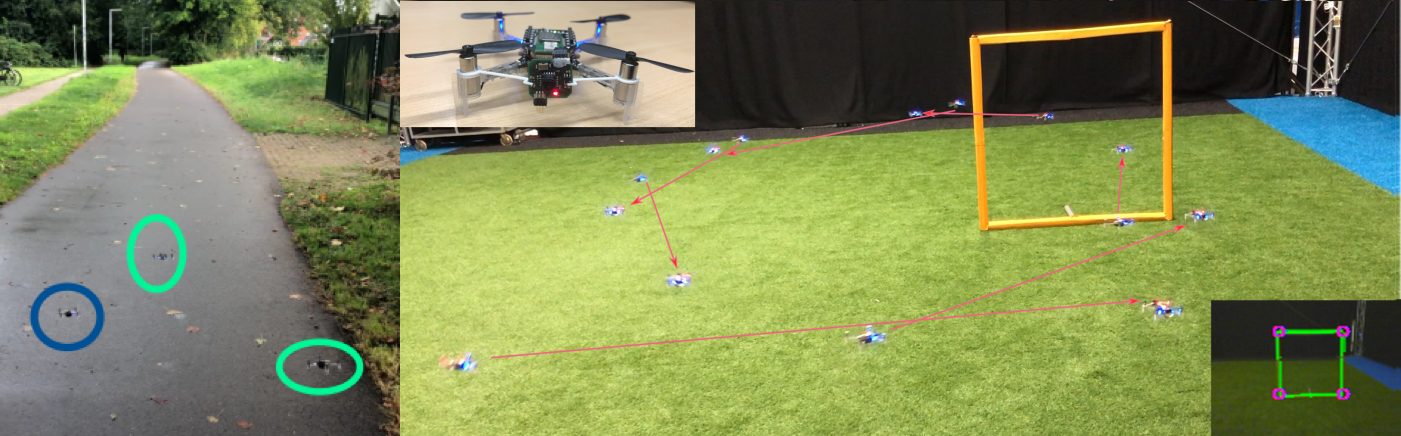}
    \caption{More flight experiments based on the relative localization. \textbf{Left}: outdoor formation flights of three Crazyflies. \textbf{Right}: leader-follower flight.}
    \label{fig:uwb_more_exp}
\end{figure}

Fig. \ref{fig:uwb_more_exp} demonstrates more experiments which all rely on the onboard relative localization, e.g., outdoor autonomous formation flight of three Crazyflies with wind disturbances, and autonomous leader-follower flight through a window based on the visual object detection on the leader and non-visual relative localization on the follower.

These experiments show that when the proposed relative localization method is used in the control loop, it has consistent convergence in practical experiments even under different unobservable states such as formation flight or hovering. Hence, it corroborates our proof on the self-regulated convergence, and the initialization convergence, just like the simulation experiments.

\section{Conclusions}
\label{sec:uwb_conclusion}

This paper proposed a fast and accurate relative localization method for fully autonomous swarms of resource-constrained robots.
First, a UWB swarm ranging protocol was proposed to support a large number of robots ranging at a high frequency.
Second, a tiny relative localization scheme was designed for resource-constrained lightweight aerial vehicles, and it is surprising to discover that the unobservability poses no issues.
Third, by experiment, less than 0.2m position error is achieved at the frequency of 16Hz for as many as 13 drones. 
This is the first fully autonomous, onboard fast and accurate relative localization scheme implemented on a team of 13 lightweight and resource-constrained aerial vehicles
The code is open-sourced, which is also readily applied to many other less resource-restricted robots.

\bibliographystyle{IEEEtran}
\bibliography{main.bib}
\end{document}